# TELUGU OCR FRAMEWORK USING DEEP LEARNING

By Rakesh Achanta[*], and Trevor Hastie[*]

*Stanford University*[*]

*Abstract*: In this paper, we address the task of Optical Character Recognition(OCR) for the Telugu script. We present an end-to-end framework that segments the text image, classifies the characters and extracts lines using a language model. The segmentation is based on mathematical morphology. The classification module, which is the most challenging task of the three, is a deep convolutional neural network. The language is modelled as a third degree markov chain at the glyph level. Telugu script is a complex *alphasyllabary* and the language is agglutinative, making the problem hard. In this paper we apply the latest advances in neural networks to achieve state-of-the-art error rates. We also review convolutional neural networks in great detail and expound the statistical justification behind the many tricks needed to make Deep Learning work.

**1. Introduction.** There has been limited study in the development of an end-to-end OCR system for the Telugu script. While the availability of a huge online corpus of scanned documents warrants the necessity for an OCR system, the complex script and agglutinative grammar make the problem hard. Building a system that works well on real-world documents containing noise and erasure is even more challenging.

The task of OCR is mainly split into segmentation and recognition. The design of each is guided by that of the other. The more robust (to noise, erasure, skew etc.) the segmentation is, the easier the task of the recognizer becomes, and vice-versa. The techniques used in segmentation are somewhat similar across scripts. This is because, usually, one connected component (a contiguous region of ink) can be extracted to give one unit of written text. While this principle applies to the Roman scripts with few exceptions; it does not hold for complex scripts like Devanagari and Arabic, where words, not letters, are written in one contiguous piece of ink. The Telugu script is of intermediate complexity, where consonant-vowel pairs are written as one unit.

The recognition task is traditionally split into feature extraction and classification. The former has been hand-engineered for a very long time. As early as 1977, Telugu OCR systems used features that encode the curves that trace a letter, and compare this encoding with a set of predefined templates (Rajasekaran and Deekshatulu, 1977; Rao and Ajitha, 1995). The first attempt to use neural networks for Telugu OCR to our knowledge was in Sukhaswami, Seetharamulu and Pujari (1995). They train multiple neural networks, and pre-classify an input image based on its aspect ratio and feed it to the corresponding network. This reduces the number of classes that each sub-network needs to learn. But this is likely to increase error rate, as failure in pre-classification is not recoverable. The neural network employed is a Hopfield net on a down-sampled vectorized image. Later work on Telugu OCR primarily followed the featurization-classification paradigm. Combinations like ink-based features with nearest class centroid (Negi, Bhagvati and Krishna, 2001); ink-gradients with nearest neighbours (Lakshmi and Patvardhan, 2002); principal components with support vector machines (Jawahar, Kumar and Kiran, 2003); wavelet features with Hopfield nets (Pujari et al., 2004) were used. More recent work in this field (Kumar et al., 2011) is centred around improving the supporting modules like segmentation, skew-correction and language modelling. While our work was under review, Google Drive added an OCR functionality which works for Telugu and a lot of other world languages. Although the details of it are not public, it seems to be based on their Tesseract multi-lingual OCR system (Smith, 2007) augmented with neural networks.

This paper improves on previous work in a few significant ways. While previous work was restricted to using only a handful of fonts, we develop a robust font-independent OCR system by using training data from fifty fonts in four styles. This data (along with the rest of the OCR program) is publicly released to act as a benchmark for future research. The training and test data are big and diverse enough to let one get reliable estimates of accuracy. Our classifier achieves near human classification rate. We also integrate a much more advanced language model, which also helps us recover broken letters. Our system performs better than the one offered by Google, to the best of our knowledge, the only other publicly available OCR for Telugu.

In our work, we break from the above mentioned 'featurize and classify' paradigm. We employ a convolutional neural network(CNN), which learns the two tasks in tandem. In addition, a CNN also exploits the correlation between adjacent pixels in the two dimensional space (LeCun et al., 1998). Originally introduced for digit classification (a sub-task of OCR), CNNs have been adapted to classify arbitrary colour images from even a thousand classes (Krizhevsky, Sutskever and Hinton, 2012). This was aided in part by better regularization techniques like training data augmentation (Simard,

Steinkraus and Platt, 2003), dropout (Hinton et al., 2012) and by increased computing power. We review the new technological "tricks" associated with Deep Learning, and apply them and some of our own in developing a CNN for our task.

The rest of the paper is organized as follows. We introduce the problem in Section 2, describing the Telugu language and its calligraphy. Section 3 explains the segmentation process and Section 4 generation of training data. Section 5 describes CNNs in complete detail, with a subsection (5.6) on our architecture. Results are discussed in 5.7 and regularization in 5.9. The language model, including the recovery of broken letters, is described in Section 6. We conclude with an evaluation of the end-to-end system in Section 7.

**2. The Problem.** Telugu is a Dravidian language with over 80 million speakers, mainly in the southern India state of Andhra Pradesh. It has a strong consonant-vowel structure, i.e. most consonants are immediately followed by a vowel. A consonant and a vowel combine to give a syllable. For example the syllable `ka` in the word `sakarma` in Figure 1 is one such entity, and `ra` in the same figure is another. Each such syllable is written as one contiguous ligature. This alphasyllabic form of writing is called an *abugida* as opposed to an *alphabet*. There are 16 vowels and 37 consonants which combine to give over 500 simple syllables. Of them about 400 are commonly used. There are nearly sixty other symbols including vowel-less consonants (`m` in Figure 1), punctuation and numbers. This brings the total number of symbols used in common writing to approximately 460. Figure 1 demonstrates the Telugu calligraphy via an example.

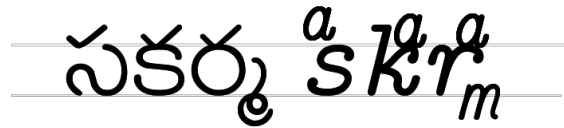

FIGURE 1. *The word* సకర్మ*(sakarma) rendered in Telugu and in English with Telugu styling. The* ✓ *corresponds to the vowel* a*, usually attached to the consonant. The syllable* స*(sa) is written in two parts,* క*(ka) in one, and* ర్మ*(rma) is split into* ర*(ra) and* ౖమ*(m).*

Our goal is to develop an end-to-end system that takes an image of Telugu text and converts it to Unicode text. This would mean that we first need to detect lines in the image. Once the lines are detected, we need to further segment each line's sub-image into individual letters. In this paper, we will refer to the extracted connected components as glyphs. Our definition of a glyph is one contiguous segment of ink, which typically maps to one syllable, like ka, ra and m in Figure 1. While a syllable can be composed of more than one connected component, like sa in the same figure, we can think of a glyph as the visual for a syllable. After a glyph is obtained, it needs to be recognized as representing one of the 460 classes. Once the candidates for each glyph (along with their probabilities) are obtained, we infer the most likely line of text from them using a language model.

**3. Segmentation.**   Given a binary image of text, where 'ink' is unity and background is zero, we first find its row-ink-marginal, i.e. the running count of number of pixels that are ON. The image is skew corrected to maximize the variance of the first differential of this marginal. That is, we find the rotation that results in the most sudden rises and falls in the row-pixel-counts (see Figure 2).

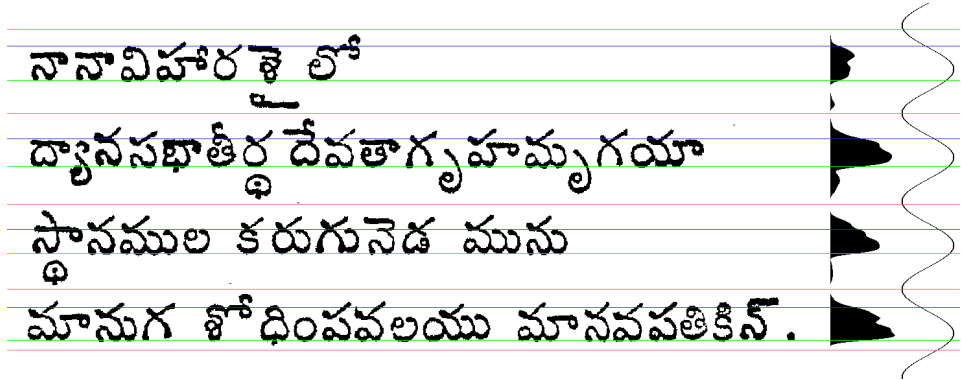

FIGURE 2. *A sample Telugu poem along with its row-ink-marginal. The detected line-separations are in red, top-lines in blue and bottom-lines in green. The smoothened row-ink-marginal and the best fitting harmonic are shown to the right. Lines are detected by looking for sudden jumps in the marginal and using the wavelength of the harmonic as a heuristic.*

Next we estimate the number of lines in the page in a novel way. This is done by taking the Discrete Fourier Transform of the mean-centred marginal. The wavelength of the harmonic with the highest amplitude gives us an estimate of the distance between two baselines. We use it as a heuristic to look for sudden drops in the row-ink-marginal and tag them as baselines. We then detect the zero between two baselines and identify them as line separations. Additionally we also identify the toplines as the point of highest gain in pixel count between the line-separation and the baseline. Thus the image is separated into sub-images of lines. Figure 2 shows an example. Notice that our algorithm is robust to descenders that are completely below the baseline.

Each line is processed to extract connected components using Leptonica (Bloomberg, 2007), an image-processing library. We thus have, as input to the classifier, the arbitrary-sized glyph and its location in reference to the base and top lines as in Figure 2. Figure 3 shows some glyphs ready to be classified.

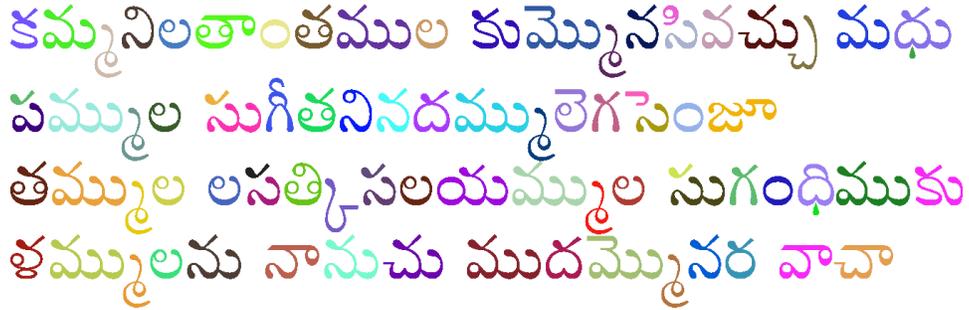

FIGURE 3. *Glyphs extracted from a text image. Each connected component, shown in a different colour, is standardized before being fed to the Neural Network.*

**4. Generating the Training Data.** As there is no publicly available training data to train a classifier with, we set out to generate our own in a mechanized and inexpensive manner. We assemble a moderate sized corpus (150 MB) of Unicode Telugu text from the internet. Based on it, we generate sample text containing all possible glyphs. This text is rendered onto digital images in four different styles (normal, **bold**, *italic* and ***bold-italic***) in fifty different fonts. The glyphs are recovered from these images via the segmentation process described above. This results in nearly 160 unique rendering per glyph. Hence in all, we have $160 \times 460 \approx 73,000$ labelled samples. Figure 4 shows all the renderings for a sample class. We also know the location information for each glyph, i.e. its position relative the top line and baselines. We expect to use this information for better classification; since the same glyph, when placed at different locations relative to the top and base lines, results in it being in different classes. Figure 5 shows an example.

Now we have the training data that can be used to train any machine learning algorithm of interest.

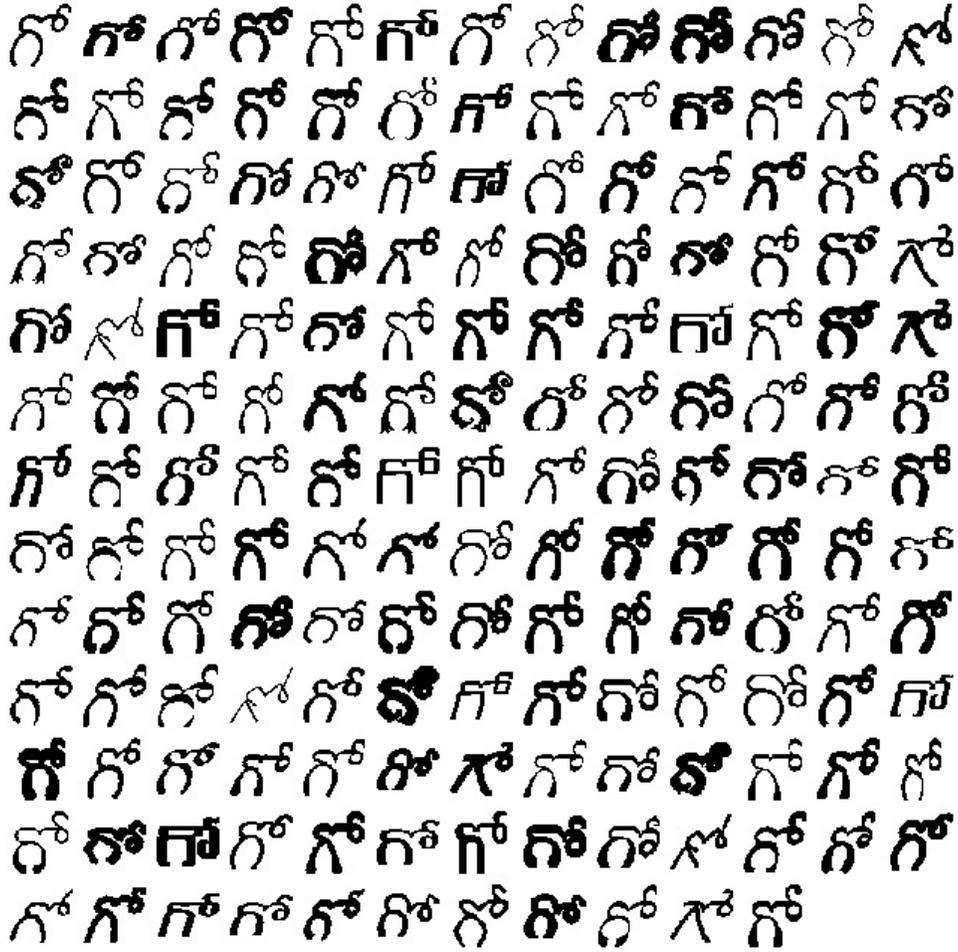

FIGURE 4. *The 167 unique renderings of the letter గో(gō). This illustrates the expected spread in the distribution of glyphs. Similarly we have nearly as many unique renderings for each of the 460 classes.*

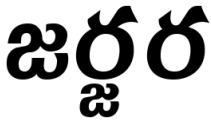

FIGURE 5. *The word jarjara. The The letter జ(ja) when rendered above the base line has its usual interpretation. But when rendered below, it loses its vowel a.*

**5. Convolutional Neural Networks.** Convolutional Neural Networks (CNNs) have been shown to be very successful for image recognition tasks over a broad spectrum of problems from digit recognition (LeCun et al., 1998) to the ImageNet classification (Krizhevsky, Sutskever and Hinton, 2012), and to the more recent Chinese hand-written character recognition task (Cireşan and Schmidhuber, 2013). This motivates us to use CNNs for the Telugu recognition task. But our problem differs from digit recognition in that it has many more classes (460 as opposed to 10). It also differs from the Chinese OCR task in that we need to incorporate location information (see Figure 5) into our classifier.

CNNs were introduced to exploit the two-dimensional correlation structure of image data. A vanilla neural network which ignores this structure (by flattening the input image) is not well suited for such pattern recognition tasks. The convolutional and pooling layers that operate on two or three dimensional image data are at the heart of the CNN architecture.

5.1. *Convolutional Layer.* In full generality, input to a CNN is a 3D-image with two spatial dimensions and one frequency dimension. For colour images, each sample is a stack of three *maps*; one each for red, blue and green. Hyper-spectral images could have many more, while gray-scale images have only one map. In our case, the input is a $48 \times 48$ single-map binary image. Given that the convolution operation employed by CNNs is very complicated, we begin by explaining a simple convolution operation on a 2D image.

*2D-Convolution.* A 2D-convolution operation is parametrised by a 2D-kernel of size, say, $k = 2l + 1$. Typically $k$ is set to three or five (however, for smaller input images, it could be set to even two). At each pixel in the interior of the input image, this kernel is applied. That is, we take the dot-product of the kernel with the $l$-neighbourhood of the target pixel. Convolution with a random kernel thus performs a local weighted-averaging over the input image. More useful kernels tend to act as various kinds of edge/feature detectors. Convolution with a $2l + 1$ kernel reduces the side of the image by $l$, which might lead to loss of information at the borders. To avoid this, the original image can be zero-padded by $l$ pixels. Figure 6 shows a few such kernels applied to a binary image.

*3D-Convolution.* A 3D image is a stack of 2D images (which are referred to as *maps* in this context). In CNNs, intermediate representations of the input image are in fact 3D images with possibly thousands of maps. These maps tend to be of a smaller size than that of the input.

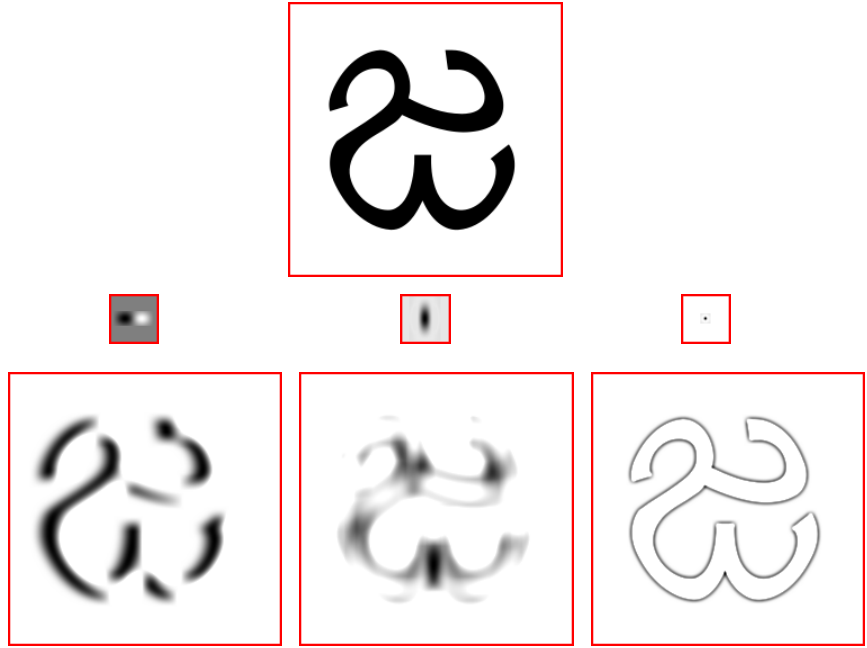

FIGURE 6. *2D Convolution: The original binary image of the letter ಇ(ja) is shown at the top, the three sample convolution kernels at the centre, and the corresponding outputs at the bottom.*

As shown in Figure 7, each constituent map of the 3D input image is convolved with its corresponding 2D kernel. The dot-products are then summed to generate a single output map. Hence an input with $d_{in}$ maps gets convolved with a 3D kernel of size $k \times k$ (and depth also $d_{in}$) to give a 'single' output map. Where $d_{out}$ output maps are required, we employ as many such kernels. Thus a 3D to 3D convolution operation has a kernel of size $d_{out} \times d_{in} \times k \times k$. In all, the 3D convolution kernel has $k^2 d_{in} d_{out}$ parameters. We can think of this operation as a series of $d_{out}$ feature extractors. The convolution operation $C'_W$ on a 3D input tensor $A$ (of size $d_{in} \times s \times s$) with a kernel $W$ (of size $k = 2l + 1$) can be written as:

$$(5.1) \qquad C'_W : \mathbb{R}^{d_{in} \times s \times s} \to \mathbb{R}^{d_{out} \times s \times s}$$

$$(5.2) \qquad C'_W(A)_{z,x,y} = \sum_{m=1}^{d_{in}} \sum_{i=-l}^{l} \sum_{j=-l}^{l} W^z_{m,i,j} A_{m,x+i,y+j}$$

with the assumption that for any indices out of range, $A$ is zero. A non-linearity $R$ is applied after the convolution $C'_W$ to give the final $C_W$ operation

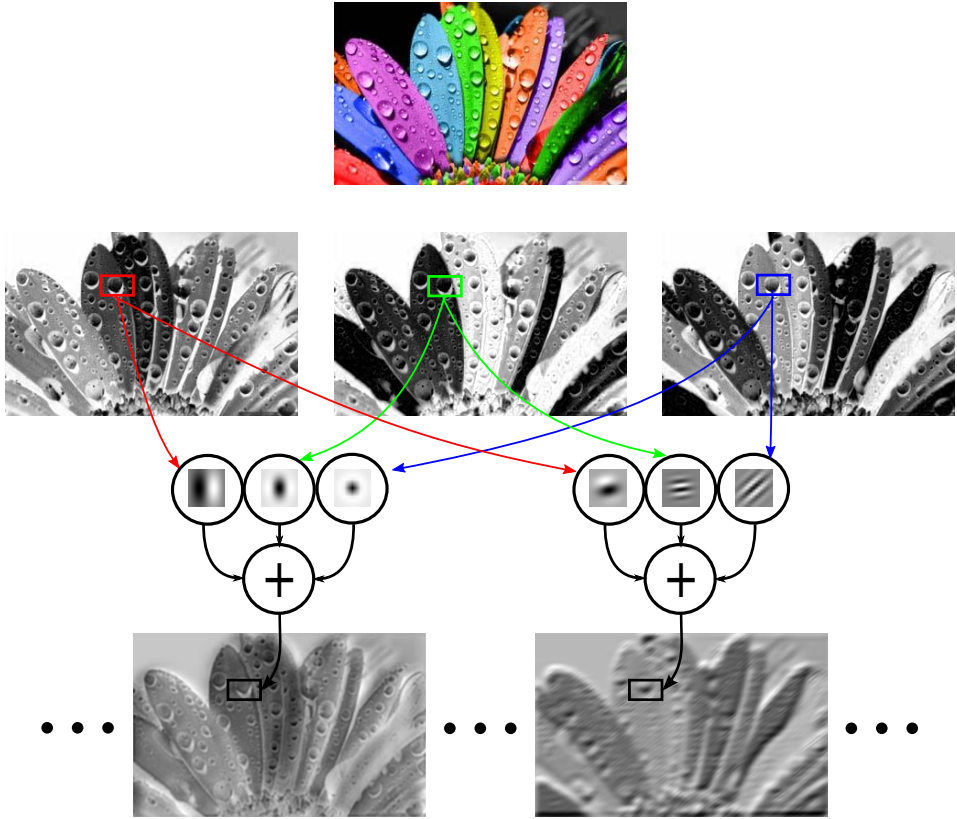

FIGURE 7. *3D Convolution: An input color image, its RBG maps, the convolutional kernels and the output maps. Many such output maps are generated to form the 3D output image. Note, however, that the maps lose the RGB interpretation after the first layer.*

of the convolutional layer.

$$(5.3) \qquad\qquad C_W(A) = R(C'_W(A))$$

Here, $W$ is the weight or parameter tensor that is to be learned by training the network. We explain the role of non-linearities shortly; for now they can be thought of as a thresholding to identify interesting features.

5.2. *Pool Layer.* A typical convolution layer outputs many more maps than it takes in. There is also high correlation between the adjacent output values in a map. Hence it makes sense to somehow 'scale' the maps down, generally by a factor of two in each co-ordinate. This operation is called pooling. Typically pooling is done over a $2 \times 2$ grid. The maximum of the four pixels in the grid is extracted as output. Taking maximum (as opposed to the average) gives the neural network translational invariance, which is key to good classification. We can increase the number of maps four-fold at each convolutional layer and decrease the area of the image-maps by a factor of four at the succeeding pool layer. This preserves the size of the image while at the same time transforming it to a different, more useful, space. The $2 \times 2$ max-pooling operation on a tensor $A$ can be written as

$$(5.4) \qquad P_2(A)_{z,x,y} = \max_{\substack{i=2x-1,2x \\ j=2y-1,2y}} A_{z,i,j}$$

5.3. *Fully-connected Layer.* CNNs can have anywhere between two to twenty convolutional and pooling layers. Their final output, which is a 3D image, is flattened into a vector. One or more fully-connected layers follow. A fully-connected layer is basically a simple matrix multiplication followed by a non-linearity.

$$(5.5) \qquad F_W : \mathbb{R}^{n_1} \to \mathbb{R}^{n_2}$$

$$(5.6) \qquad F_W(A) = R(WA)$$

Here again $W$ is the weight matrix that is to be learned by training the network, and $R$ is a general non-linearity (usually applied element-wise). Note that this is nothing but the typical hidden layer of a traditional neural network.

5.4. *Output Layer.* The last of such fully-connected layers is the output layer $F_W^S$. It has as many nodes as the number of classes, $K$, in the problem. It employs a specific form of non-linearity, the *softmax* function $S$.

$$(5.7) \qquad F_W^S : \mathbb{R}^{n_2} \to \mathbb{R}^K$$

$$(5.8) \qquad F_W^S(A) = S(WA)$$

where,

$$(5.9) \qquad S_k(A) = \frac{e^{A_k}}{\sum_{j=1}^{K} e^{A_j}}$$

This is the same transformation used in the multilogit model, and produces positive values that sum to one. The $K$-vector obtained from the matrix multiplication is exponentiated to make it all positive and then normalized to belong to the $K$-simplex. These $K$ values are interpreted as the class probabilities. In summary, one can think of the convolutional and pool layers as the feature extraction phase of the CNN and the the fully connected layers as a vanilla neural network classifier on top.

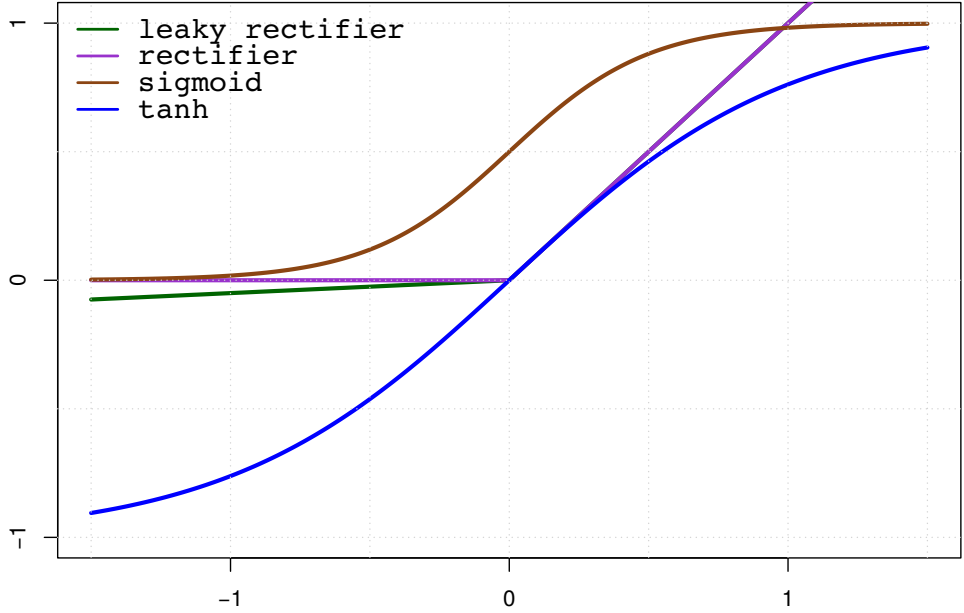

Figure 8. *The four most common neural network activations.*

Let $X, Y$ denote a random image and its class label respectively. Then, according to our model, the likelihood is given by

$$(5.10) \qquad P(Y = y \mid X = x; \mathcal{W}) = p_y(x; \mathcal{W}) = S_y(W_1 A(x))$$

where $A(x)$ is the input to the softmax layer (for a given set of network parameters $\mathcal{W} = \{W_i\}$). Neural networks are trained to maximize the log of the above likelihood over the training data $\mathcal{D}$ to find the optimum network parameters $\mathcal{W}^\star$.

$$(5.11) \qquad L = \sum_{(x,y) \in \mathcal{D}} \log\left(p_y(x; \mathcal{W})\right)$$

$$(5.12) \qquad \mathcal{W}^\star = \underset{\mathcal{W}}{\operatorname{argmax}} \, L$$

5.5. *Non-linearities.* Non-linearities are applied to the output of each layer of the neural network. A neural network with one non-linear hidden layer can act as a universal function approximator (Cybenko, 1989), i.e. it can approximate any continuous function on a compact space (given enough nodes). The Sigmoid function has had been the most commonly used non-linearity. It and the closely related hyperbolic tangent function resemble step functions.

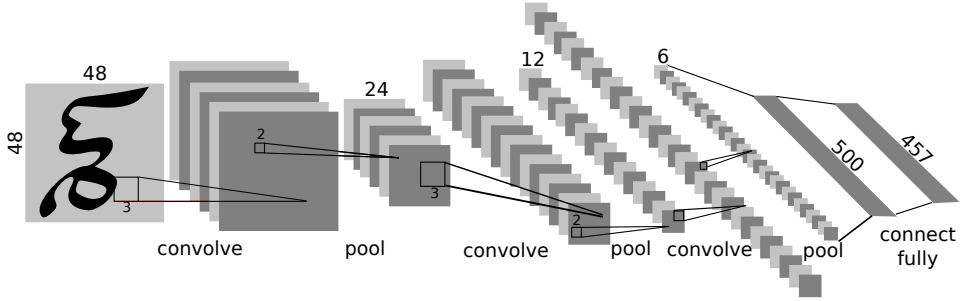

FIGURE 9. *A `traditional` convolutional neural network where a single channel 48×48 input is subjected to three convolution-pool operations (resulting in $6 \times 24 \times 24$, $16 \times 12 \times 12$, $30 \times 6 \times 6$ maps respectively). The final $30 \times 6 \times 6$ tensor is flattened to 1080 nodes and is fully connected to a hidden layer with 500 units, followed by a multi-class logistic layer over the 457 classes.*

$$(5.13) \qquad \text{sigmoid}(x) = \frac{1}{1 + e^{-x}}$$

$$(5.14) \qquad \tanh(x) = \frac{e^x - e^{-x}}{e^x + e^{-x}} = 2\,\text{sigmoid}(2x) - 1$$

Each sigmoid node in the hidden layer simulates a step function of a different size at a different location. The output node combines these steps to approximate the desired mapping. Without the non-linear activations, a neural net, regardless of depth, can only act as a linear transformer.

In recent years rectifier linear units(ReLU) have gained in popularity (Nair and Hinton, 2010). A rectifier is just the $x_+$ function, in a sense it is the most basic non-linearity.

$$(5.15) \qquad \text{ReLU}(x) = x_+ = \max(0, x)$$

Despite their simplicity and zero gradient in half of the domain, they improve performance of our network in a significant way. However, a rectifier activated network needs multiple starts (around ten in our case) to get a successful training instance. This is because of a propensity for the gradients to either blow up in the linear region or die down in the flat region. It has been observed that performance can be further improved by allowing the rectifier to 'leak' a bit. That is, instead of being zero for negative values, it has a small slope.

$$(5.16) \qquad x_- = \max(0, -x)$$

$$(5.17) \qquad \text{LeakyReLU}(x) = x_+ - \alpha x_-$$

The choice of $\alpha$ is arbitrary. Experiments to learn the amount of 'leakage' (He et al., 2015) demonstrate that it is desirable to decrease $\alpha$ as we move from input to output layers. For our network, instead of trying to learn the leakages, we fix them as follows. We start with $\alpha = .5$ for the first convolutional layer and decrease it at each layer to finally end with an $\alpha = .05$ for the last hidden layer. Figure 8 shows the various non-linearities and Table 2 summarizes the effect of these choices on classification error for our application.

5.6. *Our Neural Network.* The glyphs obtained from segmentation are scaled up or down to a $48 \times 48$ square image and supplied as input to the CNN. We preserve the aspect ratio while scaling. In addition to these 2304 binary pixel values, we have two numbers representing the locations of the top and baselines, which we later incorporate into the network. We use a `traditional` architecture, based on LeCun et al. (1998), as our reference point to compare various design choices and regularizations. It has three pairs of convolutional-pool(`conv-pool`) layers, all of which employ a $3 \times 3$ convolution kernel and a $2 \times 2$ max-pooling window. These are followed by two fully connected layers. We use leaky-ReLUs in place of the tanh activations of LeCun et al. (1998). The last layer's soft-max activation yields the final class "probabilities" $p(x; \mathcal{W})$. Using the notation developed in this section so far, this model can can be written as:

$$(5.18) \qquad p(x; \mathcal{W}) = F_{W_5}^S \circ F_{W_4} \circ P_2 \circ C_{W_3} \circ P_2 \circ C_{W_2} \circ P_2 \circ C_{W_1}(x)$$

Each term in the above composition is a layer of the CNN. $C_W$, $P_2$, $F_W$ represent convolutional, pooling and fully-connected layers respectively parametrized by weight vector $W$. $\mathcal{W} \equiv \{W_i\}$ is the set of CNN parameters to be learned. Figure 9 shows the traditional model. Figure 10 shows the output at each layer for a sample input. Figure 11 shows intermediate and final features for some sample inputs.

5.7. *Discussion of Results.* Table 1 summarizes error rates, sizes and speeds of different architectures. The k-Nearest Neighbour classifier does very poorly on this data with a 35% test error. This justifies the need to use more sophisticated models. With multi-class logistic regression, the test error comes down to 32%. It further reduces to 27% with a simple linear SVM, which is a more robust classifier. Both of these have around a million parameters, which can be considered as a point of reference for model size. A conventional neural network with two hidden layers, each with two thousand nodes, takes the error down to 13% while using nearly 10 million parameters. All these architectures ignore the 2D structure of the input data and are not specialized for the problem at hand. A structure-exploiting CNN with three conv-pool operations followed by two fully connected layers gives an error rate of 1.48% with nearly as many parameters as a multi-class logistic model. Deeper networks with more convolutional layers and only one fully connected layer give better and near human accuracies for this problem, while keeping the network size small.

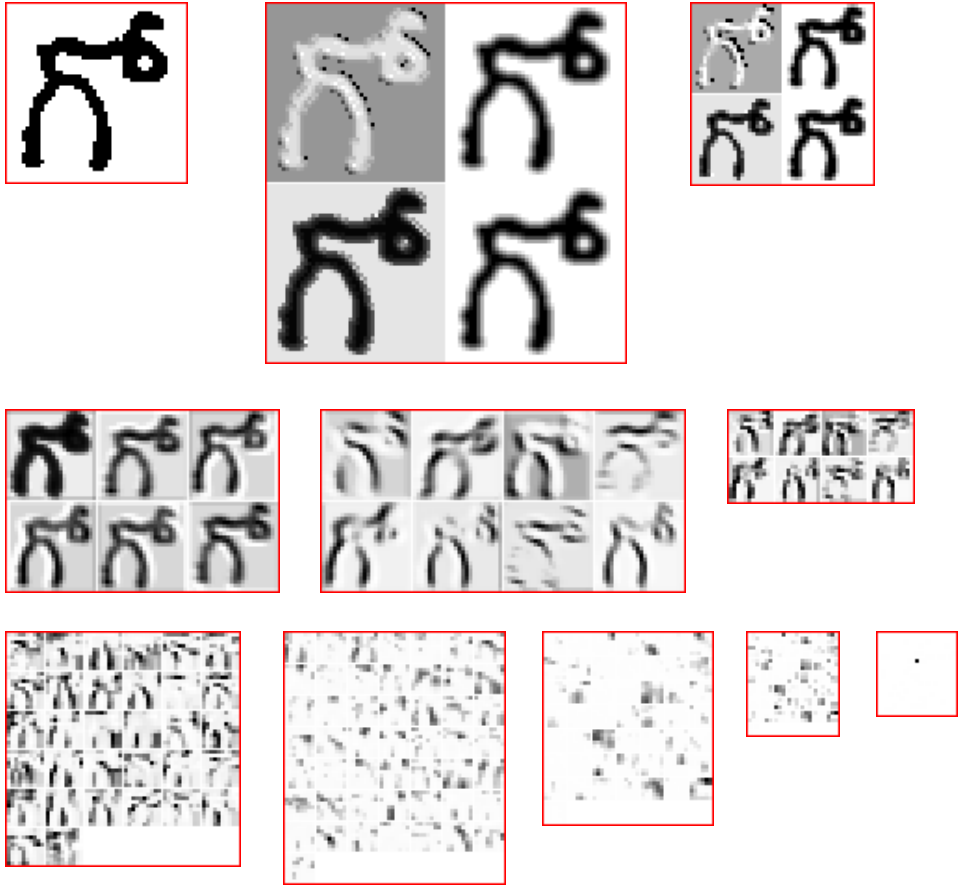

FIGURE 10. *The letter ङ (gṏ) goes through the 'slim' network from Table 1. The first row shows the input, a convolution and a pooling (48x48-4C3-MP2). The second, two convolutions and a pooling (6C3-8C3-MP2). The third, three convolutions followed by a pooling and a softmax layer (32C3-50C3-50C3-MP2-457SM). The last two images are hence the input and output to the softmax classification layer, with the single dark dot indicating a very high probability for the class ङ (gṏ).*

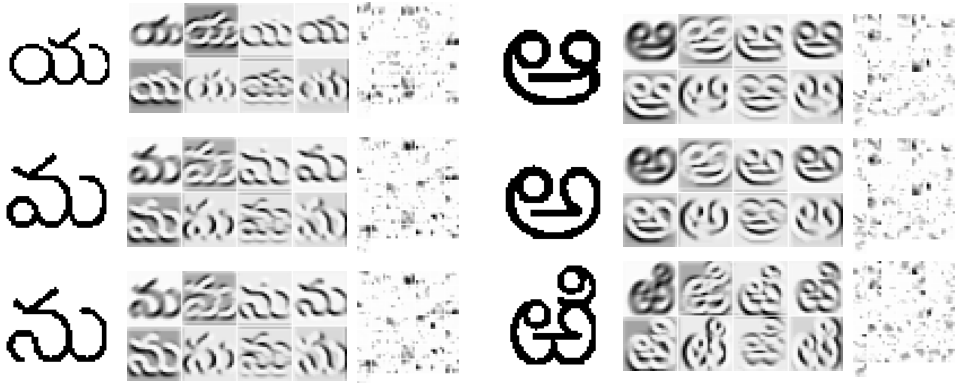

FIGURE 11. *Intermediate features and final transformations for various input images (extracted from the fourth and the last hidden layers respectively of the 'slim' network). Notice how similar looking images have similar representations in the transformed space.*

One can make the network size as small as 100K parameters while still achieving test error rates below 1.5%. For these deeper networks (with only the one fully connected layer) half of the parameters are in this last 'classification' layer. The rest of the network with its six to eight convolutional layers uses the other half of the parameters. Thus the network parameters are equally split between the classification and feature-extraction tasks. However, the convolution operation is computationally expensive and can significantly increase the time required to classify a glyph. But as we are in the sub millisecond range, this is acceptable. Even the deepest neural networks, once trained, classify much faster than a kNN classifier. Also, in the context of the larger Telugu OCR problem, the Viterbi algorithm employed by the language model is the main speed bottle neck.

TABLE 1

*Performance of various architectures. All networks employ input distortion as regularization. $48 \times 48$ represents the binary input image, $m$C3 a $3 \times 3$ convolutional layer with $m$ output maps, MP2 a $2 \times 2$ max-pooling layer, $m$N a fully connected layer with $m$ units and 457SM a softmax output layer on 457 classes. Speed is specified in terms of time taken to classify one image on a GPU. Network size is specified in millions of parameters. The standard error of the error estimates is $\approx .1\%$.*

|   | Architecture | %Training Error | %Test Error | Speed (ms) | Size (M) |
|---|---|---|---|---|---|
| 1 | Ten nearest neighbour classifier | 26.40 | 35.55 | $\gg$10 | – |
| 2 | Linear support vector machine | 25.16 | 28.65 | .023 | 1.05 |
| 3 | Multi-class logistic (48x48-457SM) | 29.08 | 32.06 | .039 | 1.05 |
| 4 | One Hidden Layer (48x48-1000N-457SM) | 24.63 | 28.38 | .057 | 2.76 |
| 5 | 48x48-2000N-2000N-457SM | 09.20 | 13.00 | .130 | 9.52 |
| 6 | 48x48-8C3-MP2-24C3-MP2-72C3-MP2-500N-457SM (traditional) | 00.58 | 01.48 | .259 | 1.14 |
| 7 | 48x48-6C3-6C3-MP2-18C3-18C3-MP2-54C3-54C3-MP2-162C3-162C3-MP2-457SM (deepest) | 00.09 | 00.74 | .646 | 0.75 |
| 8 | 48x48-4C3-MP2-6C3-8C3-MP2-32C3-50C3-50C3-MP2-457SM (slim) | 00.41 | 01.36 | .168 | 0.25 |

5.8. *Design Choices.* The fourteen layered neural network from Table 1 (line 7) requires the specification of nearly a hundred hyper-parameters. It does take quite a bit of experimentation to obtain the set of hyper-parameters that lead to good prediction. The 'slim' architecture from Table 1 has been carefully hand-crafted to improve both on size and speed over the `traditional` architecture. One could spend countless hours tuning these hyper-parameters. This difficulty in picking the right architecture and hyper-parameters gives neural networks an air of mystery. Here we will briefly point out some of the important design choices in an attempt to demystify the process.

*Leaky ReLU.* In our dataset of binary images, nearly one-fifth of the pixels are ink and the rest are background. If we consider ink to be one and background zero, the mean of the image is at 20%. One can instead consider the reverse and increase the mean to 80%. A higher mean will help keep the ReLU units activated (i.e. in their linear, as opposed to flat, region). This is a non-issue for the leaky ReLUs, which are never really flat. In addition to more successful training instances, the leaky ReLUs also improve performance by about 25%. Using traditional tanh activation not only increases the computational complexity but also reduces performance by 25%. Table 2 summarizes these effects.

Table 2

*The effect of various design choices on test and training errors. We use the `traditional` architecture from line 6 of Table 1. We report median rates over eleven training attempts. The standard error of the error estimates is $\approx .1\%$*

| Design Choices | %Train Error | %Test Error |
|---|---|---|
| `traditional` architecture | 0.58 | 1.48 |
| No input inversion | 0.56 | 1.40 |
| No leaky `ReLUs` (needs multiple starts) | 0.97 | 2.00 |
| `tanh` activations | 1.10 | 2.04 |

*Kernel Size.* The traditional CNN architecture, based on LeCun et al. (1998), has a pool layer after each convolutional layer. The kernel size decreases as the size of the image decreases with pooling. Modern architectures tend to use $3 \times 3$ kernels all through. Where bigger kernels are needed, multiple convolutional layers are used instead. For e.g., a pair of back to back convolutional layers, with kernel sizes of three each, together result in a receptive field of size five. They need $2 \times 3 \times 3 = 18$ weights in all, where a convolutional layer with a kernel size of five needs $5 \times 5 = 25$ weights, and has one less non-linearity. However, we pay the price with increased computational complexity.

5.9. *Regularization.* Given that we are fitting a $\{0,1\}^{48 \times 48} \to (0,1)^{457}$ model with over a million parameters using only 55000 training samples, it is very easy to overfit, making regularization a key component of the training process. Two forms of regularization seem to work well for this problem – dropout (Hinton et al., 2012) and input distortion. Both of them distort the input to a classifier, albeit in very different ways.

*Input distortion.* Corrupting input data to stabilize predictions has been employed by various machine learning techniques (Abu-Mostafa, 1990; Simard et al., 1998). In addition, it can be shown that that adding gaussian noise to the data results in an $L_2$-type regularization (Bishop, 1995). We go much farther than gaussian noise and apply the following distortions to the input image (all of them by a random amount): translation, elastic deformation (Simard, Steinkraus and Platt, 2003), zoom, rotation and salt & pepper noise. Although the original training data better approximates both the test and real-world data, we notice that distortion proves to be an effective form of regularization. The network is less prone to over-fitting given that it does not see the same exact sample twice. It is forced to learn invariant features instead of memorizing image patches. Figure 12 shows some sample

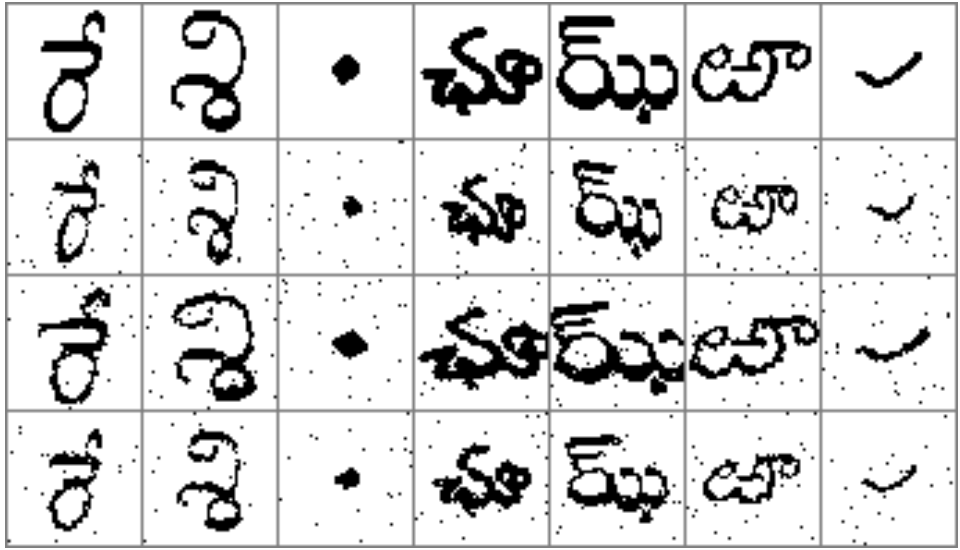

FIGURE 12. *Three distortions applied to the seven original images in the top row. Each distortion is a combination of a random amount of translation, rotation, zoom, elastic deformation and noise.*

distortions.

*Dropout.* Dropout employs multiplicative noise, where each input is either doubled or set to zero with equal probability. (More generally, an input is set to zero with a probability of $\delta$ else it is multiplied by a factor of $\frac{1}{1-\delta}$. Unless explicitly specified, $\delta$ is 1/2.) This multiplicative noise results in a penalty that is data dependent (Wager, Wang and Liang, 2013), thus incorporating the distribution of input data into the regularization process. An alternate view of dropout — the one that inspired its invention (Hinton et al., 2012) — is that we encourage the network to learn features that are useful by themselves rather than features that are useful in the presence of one another. By dropping out a random half of the features at each gradient descent step, we are preventing co-adaptation of features. In other words, consider the simple case of linear regression on two highly correlated predictors. The coefficients can co-adapt to minimize training error, but for generalization only their sum matters. Dropout prevents this by not always updating the two coefficients together, thus breaking their correlation. This helps generalization. Yet another intuition is that we are very weakly training a huge number of models and averaging them at test time.

While dropout can be applied at any layer of the CNN, it does not make much sense to use dropout at the convolution layers. If dropout is applied at the input layer, it is the same as salt & pepper noise. Dropout works best when applied to the finally extracted features just before feeding them to the softmax layer. We hypothesize that dropping out a feature (that has been extracted by eight convolutional layers) is equivalent to distorting the input image significantly. Thus dropout and input distortions have similar effects. This is also apparent in Table 3, which summarizes performance using various forms of regularization.

TABLE 3

*The effect of various forms of regularization on test and training errors. We take the* `traditional` *architecture from Table 1 and remove one form of regularization from it to see how the absence affects performance. Median rates over 11 trails are reported. The standard error of the error estimates is* $\approx .1\%$

| Regularization Variations | %Train Error | %Test Error |
|---|---|---|
| `traditional` architecture as-is | 0.58 | 1.48 |
| w/o input distortion | 0.01 | 1.77 |
| w/o dropout | 0.23 | **1.13** |
| w/o input distortion & dropout | 0.01 | 2.78 |
| w/o depth (third conv-pool layer) | 1.19 | 2.69 |
| w/o input distortion, dropout & depth | 0.00 | 4.10 |
| w/o salt & pepper noise | 0.29 | **1.02** |
| w/o just the elastic aspect of distortion | 0.48 | 1.16 |

The `traditional` architecture from line 6 of Table 1 gives a test error of 1.48%. This employs both dropout and distortion. Without both of them, the test error shoots up to 2.78%. Using just dropout it is 1.77%, using just the distortions it is 1.13%. While this might seem strange that dropout affects performance adversely in the presence of distortions, it is only because together the two forms of regularization are too much. This motivates us to examine the effect of using a lower dropout rate. Figure 13 summarizes these results. In the absence of distortions, a dropout rate of around 50% works best; in its presence little to no dropout is needed. Without salt & pepper noise we obtain better error rate of 1%, but this leaves the model susceptible to data-drift, as we expect to see some stray pixels in real-world data.

*$L_\infty$ regularization.*   Given the depth of the CNN, the gradient of the loss with respect to the parameters of the initial layers tends to be very high, even as high as $10^{16}$. This phenomenon is dubbed *gradient explosion.* This is because of the exponentially many paths from the initial layers to the output. It thus makes sense to impose an $L_\infty$ constraint on these weights, or equivalently on their gradients. This form of regularization leads to more successful training attempts without adversely affecting generalization.

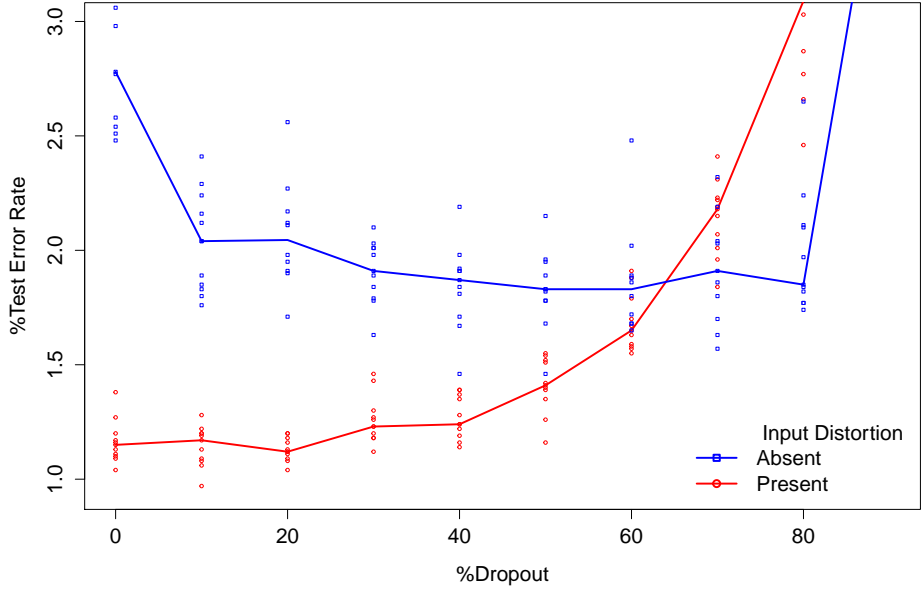

Figure 13. *The effect of dropout on test error. Each dot represents one training attempt. Solid lines show median performance for a given dropout rate over eleven such attempts. In the presence of input distortion, no to little dropout is required. In its absence, however, 40-80% helps.*

5.10. *Location Information.* Recall that when we extract a glyph's image, we also know its position relative to the top and baselines (Figure 5). Feeding location information into the neural network lowers the error-rate of our `traditional` model from 1.41% to 0.81%. This is a significant gain of 42%. This gain is mostly from glyphs like æ(ja) in Figure 5 which change meaning depending where they are located with respect to the baseline. However, there is no straight-forward way to augment the input image with the two real numbers representing the relative position of the top and baselines. In fact, these two numbers are features that do not need further processing. Therefore, we treat them as such and append them to the input of the output layer (the output layer accepts a flattened version of whatever the layer before it gives out). We should ensure that these values are at the same scale as the rest of the features that are being fed to the softmax layer.

Else, they might fail to assert themselves since we apply some sort of $L_2$ regularization on the weights (via, dropout, distortions or early-stopping). Table 4 shows performance gained by using location information.

Table 4

*Improved test error with location information for networks from Table 1*

| Architecture | %Without | %With |
|---|---|---|
| Multi-class logistic | 28.43 | 22.39 |
| tanh activated `traditional` | 02.04 | 01.64 |
| `traditional` | 01.42 | 00.82 |
| `deepest` | 00.74 | 00.56 |
| `slim` | 01.36 | 00.93 |

5.11. *Implementation and Training.*  Given an image, the full model as specified by (5.18) gives the output probabilities. We can then calculate the log-likelihood over the training data according to (5.11). The negative of the log-likelihood(NLL) is the cost we wish to minimize along with a regularization penalty, $\Phi$, on the weights. $\Phi$ is further augmented by sophisticated $L_2$-equivalent penalties like dropout, input distortion and early stopping that are incorporated via the training process.

$$(5.19) \qquad cost = -\sum_{(x,y)\in\mathcal{D}} log\left(p_y(x;\mathcal{W})\right) + \Phi(\mathcal{W})$$

While one could use the usual back-propagation algorithm to find gradients of the cost with respect to the network parameters, we use the more modern approach of symbolic differentiation. The network parameters $\mathcal{W}$ are initialized as *symbols* in the Python based software package Theano (Bastien et al., 2012; Bergstra et al., 2010). The final cost is specified in terms of these network parameters as given by the above equation. Theano gives us the gradient of the cost with respect to the network parameter symbols. Traditionally back-propagation was used to efficiently apply the chain-rule, one layer at a time, to get the desired gradients. Symbolic differentiation packages apply the chain-rule internally to give us a gradient function-object that can be evaluated at a point of interest in the $\mathcal{W}$ space. In addition to liberating us from having to write complicated back-propagating code, Theano also leverages processing power of Graphics Processor Units (GPU) available on most modern computers. It can also optimize the gradient function-object via algebraic simplifications where possible.

Once we have the gradient of our final cost with respect to the network parameters, we can perform stochastic gradient descent in the parameter space. We use mini-batch stochastic gradient descent where gradients are averaged over a 'mini-batch' of 20 training samples. We also use momentum or Nesterov averaging, where we step along a direction that is the exponentially weighted running average of recent gradients (rather than along the current gradient direction alone). The randomly initialized neural network trains for two hours, going over the entire dataset 100 times, on a system with Nvidia GTX Titan Black GPU, to give a test error rate of $\approx 1.4\%$. Training is five times slower without the GPUs. Convergence is faster with leaky ReLUs than with ReLUs, as the former have gradients of larger magnitudes (on an average). For the same reason, convergence is faster with ReLUs than with tanh activations.

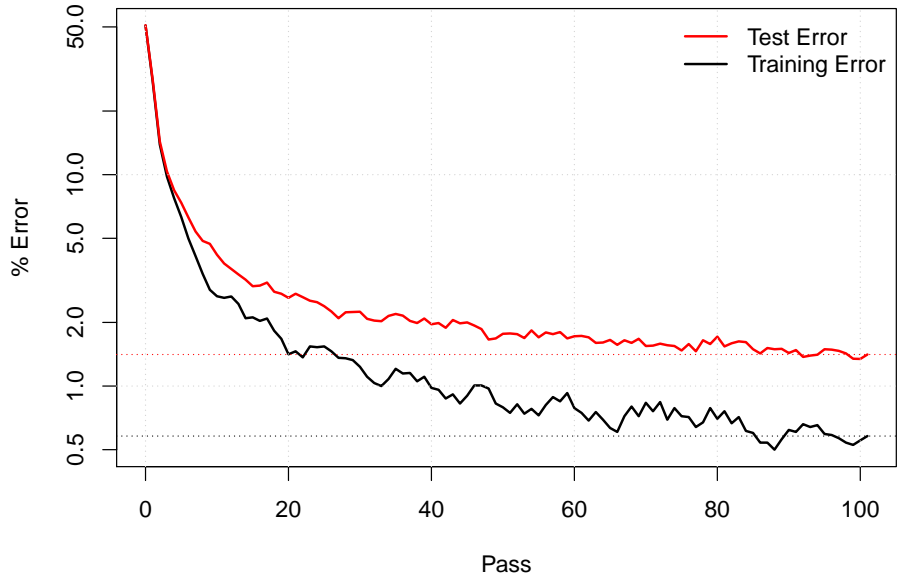

FIGURE 14. *Training and Test error as a function of passes through the data.*

**6. Language Model.** Once we have the list of probable candidates for each extracted glyph, it remains for us to find the most likely line of text. One could just consider the most likely candidate for each glyph and output the sequence of such best-matches as the final extracted text. This might not be the best thing to do, as there is a very strong dependency structure in human speech and text. For this reason, while the CNN has a very good accuracy on the training and test data, we might be able to further improve on it by using a language model to aid classification. Using a language model helps resolve any ambiguities that the CNN has between similar looking glyphs. Spurious ink can attach two glyphs together while erasure can cut a glyph into pieces. A language model can help us address both these problems. Further, we can incorporate biases into the OCR system according to the frequency of occurrence of various glyphs, letters, words, etc.

Given a sequence of input images $\mathbf{x} = (x_1, x_2, \ldots, x_t)$ where $x_i \in \{0, 1\}^{48 \times 48}$, we need to find $\mathbf{y} = (y_1, y_2, \ldots, y_t)$, the sequence of output labels that maximizes $P(\mathbf{y}|\mathbf{x})$. An estimate for $P(\mathbf{y}|\mathbf{x})$ is given by the simple formula

$$(6.1) \qquad \tilde{P}(\mathbf{y}|\mathbf{x}) \propto \tilde{P}(\mathbf{y}) \prod_i \hat{P}(y_i|x_i)$$

Here, $\tilde{P}(\mathbf{y})$ is an estimate for $P(\mathbf{y})$ (the probability of a sequence of glyphs as a sentence in the language) and $\hat{P}(y_i|x_i)$ is the probability estimate given out by our classifier when shown the image $x_i$. Intuitively, this formula reflects the human reading mechanism, where the reader combines the likelihood of a sequence of letters in the language $P(\mathbf{y})$ with what each glyph looks like $P(y_i|x_i)$. It might be surprising to note that there is no $P(y_i)$ in the above formula. It drops out because we chose to train the neural network with the same number of samples per class (even though some classes are more than a thousand times as likely as some others). A detailed derivation of the above formula is given now. We start with Bayes' rule for $P(\mathbf{y}|\mathbf{x})$.

$$(6.2) \qquad P(\mathbf{y}|\mathbf{x}) \propto P(\mathbf{x}|\mathbf{y})P(\mathbf{y})$$

First consider the term $P(\mathbf{x}|\mathbf{y})$. In reality there is a font based dependence across glyph renderings $\{x_i\}$ given corresponding labels $\{y_i\}$. But we make the independence assumption that

$$(6.3) \qquad P(\mathbf{x}|\mathbf{y}) = \prod_i P(x_i|y_i).$$

While simplifying the math, it only makes the problem harder. Additionally, it is applicable to our case as we do not incorporate font based dependencies across glyphs in our neural network (say by maintaining a state as we move from glyph to glyph). Using Bayes' rule,

$$(6.4) \qquad P(x_i|y_i) = \frac{P(y_i|x_i)P(x_i)}{P(y_i)}$$

$P(x_i)$ drops out as it is does not depend on $\mathbf{y}$. By substituting (6.3), (6.4) in (6.2), we have,

$$(6.5) \qquad P(\mathbf{y}|\mathbf{x}) \propto P(\mathbf{y}) \prod_i \frac{P(y_i|x_i)}{P(y_i)}$$

$P(y_i)$ is the unconditional probability of a glyph appearing in the language. $P(y_i|x_i)$ is the probability of a class given an image. We do not know $P(y_i|x_i)$ but can approximate it by $\hat{P}(y_i|x_i)$, which is the probability given out by our classifier when shown the image $x_i$. These class probability estimates $\hat{P}(y_i|x_i)$, however, are learned by showing the network a different distribution of $P(y_i)$, denoted by $\hat{P}(y_i)$. The network's estimate of probabilities $\hat{P}(y_i|x_i)$ for a given image are biased by $\hat{P}(y_i)$. Hence we need to perform case-control correction to get the estimate $\tilde{P}(y_i|x_i)$. Assuming $P(y_i)$ is known, we have

$$(6.6) \qquad \tilde{P}(y_i|x_i) \propto \hat{P}(y_i|x_i) \frac{P(y_i)}{\hat{P}(y_i)}$$

In our case, the training data contains equal number of instances per class. Therefore, $\hat{P}(y_i) = 1/457$ is the same for all $i$, hence it drops off from (6.6). Now, plugging in $\tilde{P}(y_i|x_i)$ as an estimate for $P(y_i|x_i)$ in (6.5), we obtain this simple formula for $\tilde{P}(\mathbf{y})$ as an estimate for $P(\mathbf{y}|\mathbf{x})$:

$$(6.7) \qquad \tilde{P}(\mathbf{y}|\mathbf{x}) \propto \tilde{P}(\mathbf{y}) \prod_i \hat{P}(y_i|x_i)$$

Once we find $\tilde{P}(\mathbf{y})$, using the $n$-gram model detailed below, we find the most probable sequence of glyphs as $\hat{\mathbf{y}} = \mathrm{argmax}_{\mathbf{y}} \tilde{P}(\mathbf{y}|\mathbf{x})$.

6.1. *n-gram Model.* Although humans employ much more complicated language models to infer from text, for the sake of character recognition we will use a simple n-gram to model $P(\mathbf{y})$. An n-gram model is an $(n-1)^{\text{th}}$ order Markov model. That is, the current unit of language is independent of all but the previous $n-1$ such units. A unigram model calculates the probability of a sequence of units as the product of their individual probabilities. But this seldom applies since we know, for example, a $q$ is most likely followed by a $u$. In English, lack of agglutination makes it possible to look up words, making them the natural choice for language units. Telugu however, is agglutinative. The sample expression *from-in-between-those-two-persons* should grammatically be written as one "word" (without the dashes). This lack of a solid sense of word identity is exacerbated by a further lack of standard convention for writing such long "words". The expressions *from-in-between those-two-persons*, *from in-between those-two-persons* are also accepted in modern verse. In addition, as the language is strongly phonetic, different accents of the same word should be rendered differently. Hence word level dictionaries are of limited use. We use a glyph as a basic language unit.

The probability of a sentence is given thus by

$$(6.8) \qquad P(\mathbf{y}) = \prod_{i=1}^{t} P(y_i | y_{i-1}, y_{i-2}, \ldots, y_{i-(n-1)})$$

Where $t$ is the number of glyphs in the sequence $\mathbf{y}$. Here, by notation, $y_0$ represents the beginning of a sentence and we ignore all $y_i$ with a negative index. In our problem, we employ a trigram (second order markov) model at the glyph level. Since a trigram is supported by $457^3 \approx 10^8$ points, it is rich enough to stand in for a dictionary. Given the syllabic nature of writing the language, it is impossible for some glyphs to occur after some other glyphs. This renders the trigram incidence matrix sparse. We take the top five candidates for $y_i$ given a glyph $x_i$ with corresponding scores $\hat{P}(y_i | x_i)$ and run the Viterbi algorithm to get the most likely sequence of glyphs according to the formula

$$(6.9) \qquad \tilde{P}(\mathbf{y}|\mathbf{x}) \propto \prod_{i=1}^{t} \tilde{P}(y_i | y_{i-1}, y_{i-2}) \hat{P}(y_i | x_i)$$

which is obtained by substituting estimates for (6.8) in (6.1) with $n = 3$. The estimates $\tilde{P}(y_i | y_{i-1}, y_{i-2})$ are learned from a digital corpus of Telugu text.

6.2. *Heuristic Over-segmentation.* A major problem in scanned text is erasure of ink. This results in glyphs being cut into two, like the *m* in Figure 15 appearing to be *rn*. Our neural network is trained only to recognize whole glyphs. Even if a glyph is cut into parts, the neural network still tries to classify the parts as whole glyphs. In this section we explain how to broaden the scope of our problem to include text where some of the glyphs are broken into two or more parts. The techniques presented in this section can be easily applied to extending the scope of the problem in the other direction, i.e., to recover glyphs that are joined together.

Our original segmentation algorithm works by looking for connected components. Two pixels are said to be connected if one belongs to the four-neighbourhood of the other. One could also consider eight neighbours instead of four. Four-connection results in more broken glyphs than eight-connection. But the latter has a higher propensity to mistakenly join two glyphs. We employ four connection and try to rejoin broken glyphs. This approach is called heuristic over-segmentation (LeCun et al., 1998). The segmentation module outputs a series of glyphs ordered from left to right (and from top to bottom when the glyphs are approximately at the same vertical location). These glyphs can be represented as a linear graph, where each edge is a glyph. A set of $n$ glyphs is a linear graph with $n + 1$ nodes. Figure 15 shows an example.

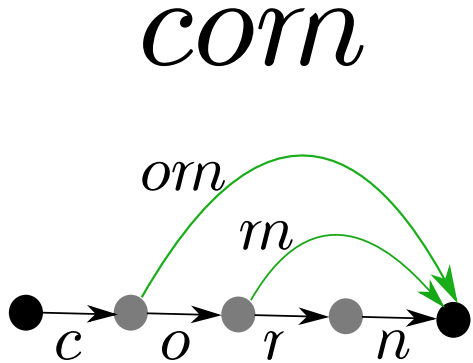

FIGURE 15. *Segmentation graph for a word that could be corn or com. Starting with the black edges, given by the segmentation module, the green edges are to be added by our recovery mechanism.*

*Combining broken glyphs.* The initial graph is linear and hence each node has at most one child. The graph is parsed bottom up, starting from the last node. At a given node, we consider the two consecutive edges from self to child and from child to grandchild. We check to see if the two edges need to be combined. This is repeated over all grandchildren of all children. Although each node has only one child to begin with, each time we decide to add a new edge, a grandchild of the current node becomes a child (like $r$ and $n$ combining to give $m$ in Figure 15). This way more than two pieces can be combined to give a candidate glyph (like $o$, $r$, $n$ combining to give $om$ in the same figure).

We currently use a set of ten heuristics to see if we should combine two glyph-pieces. These fall under the following broad categories:

- the glyphs are small
- the bounding boxes overlap
- the classifier is not confident about the glyphs
- top-matches together have low bigram probability

We assign a weight to each rule, and if the total score exceeds a threshold, we combine. These heuristics have been developed by looking at various failure cases in sample texts. Alternately, one could use a sparse logistic regression model and a large set of heuristically developed rules to get a subset of important rules (and their corresponding weights). This needs additional training data. It is better to over-combine than under, since both the original pieces and their combination are considered for further processing. In our example, both *rn* and *m* are considered by the final Viterbi decoder.

6.3. *Viterbi Decoder.* Once we have our final segmentation graph, which is a *Directed Acyclic Graph*(DAG), we generate the *recognition graph* (Le-Cun et al., 1998). For each arc in the segmentation graph, we take the corresponding image and pass it through our classifier, and get the top $M$ candidates (and their probabilities). We build a new graph by replacing each arc in the segmentation graph by $M$ weighted arcs each carrying a different candidate-character. It seems that $M = 5$ is sufficient in practice. Figure 16 shows an example with $M = 3$ matches per image.

We now need to find the path in the recognition graph that corresponds to the highest probability as defined by (6.9). Note that (6.9) has two terms per glyph: one is the probability of a candidate label given the image the other is the $n$-gram probability of this candidate label given the previous $n - 1$. The former is incorporated in the recognition graph, however the latter is not. Finding the strongest path in the recognition graph would correspond to picking the top-match for each glyph. Doing so would not incorporate any linguistic information. Hence we need to pick among all the paths in the DAG, the one that has the highest path probability as defined by (6.9). This could be computationally expensive. Suppose we have a string of thirty glyphs with five matches per glyph, we will have $5^{30}$ (over $10^{20}$) paths to consider. To solve this seemingly intractable problem, we use the famous Viterbi algorithm (Viterbi, 1967) — an efficient dynamic programming algorithm that finds the shortest path in a graph.

Before we can use the Viterbi algorithm, we need to augment the recognition graph with $n$-gram probabilities. The length of an edge is not just the likelihood of a glyph on it, instead, we need to multiply it by the $n$-gram probability of the current character-candidate given the candidates for the previous $n-1$. Each of the $M$ edges (between a node and one of its children) in the recognition graph is replaced by $M^{n-1}$ edges, where $n$ is the order of the $n$-gram. Thus, now, between a node and one of its children, we have $M^n$ edges. We call this the *n-gram graph* (Figure 17). We run the Viterbi algorithm on this graph to get the most likely path.

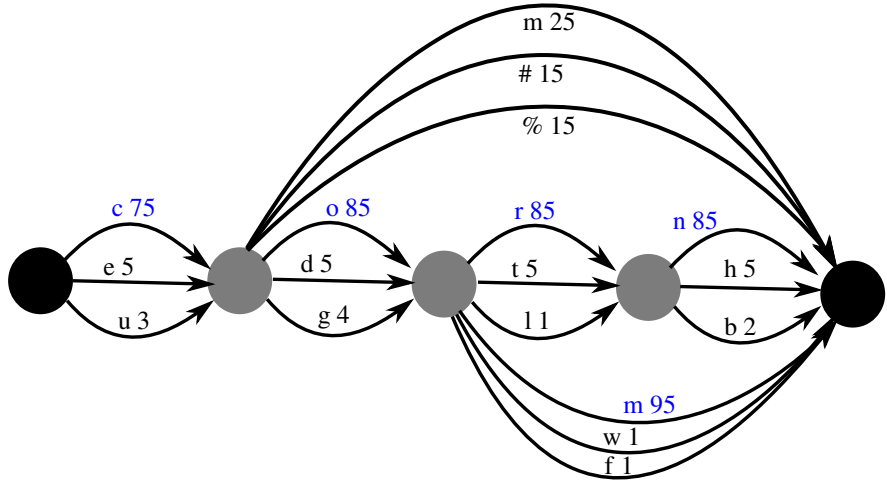

FIGURE 16. *Recognition graph for the image in Figure 15. Each arc is replaced by three arcs corresponding to the top three matches for the image (M = 3). Corresponding percentage probabilities are also shown as edge weights. The top-matches are shown in blue.*

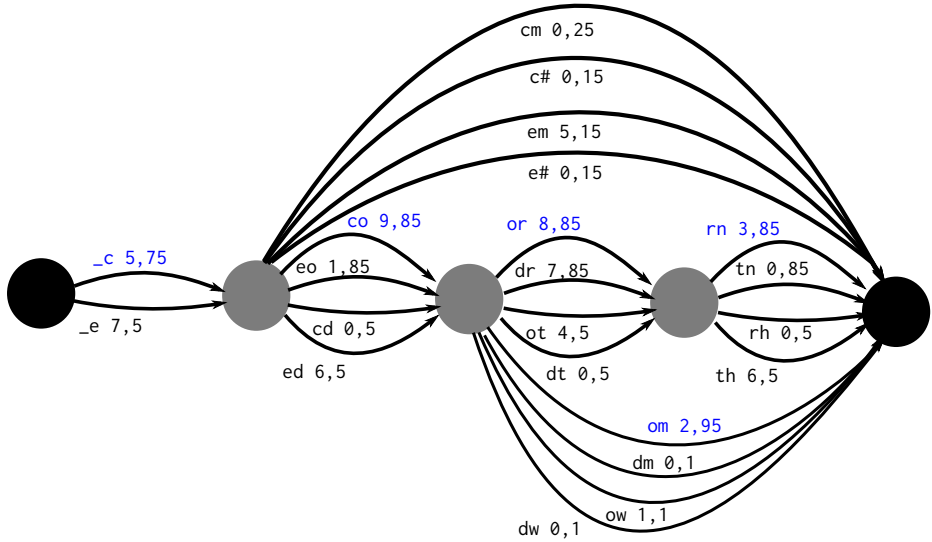

FIGURE 17. *n-gram graph, shown here with two matches per image (M = 2) and bigram probabilities (n = 2). For our application, we use M = 5, n = 3, leading to a much more complicated graph. Viterbi algorithm finds the strongest path in this graph. The* `c`*,*`o`* in the edge-label* `co 9,85`* denote* `c`* as a candidate from the previous image and* `o`* from the current.* `85`* is the %probability of* `o`* given the image; and* `9`* is the %probability of* `o`* given* `c`*. The edge strength is .09 ∗ .85 = .0765. `_` represents beginning of a sentence. Here* `_c-co-or-rn`* is the strongest path making corn the final transcription.*

6.4. *Recalibrating the probabilities.* Recall that to get an estimate of $P(y_i|x_i)$, we correct the bias of $\hat{P}(y_i|x_i)$ — the probability given by our classifier. While under the trained CNN model, $\hat{P}(y_i|x_i)$ is asymptotically consistent for $P(y_i|x_i)$, there are a few considerations:

- the CNN is only an approximation to the true model
- the choice of a particular CNN configuration (as specified by hundreds of hyper-parameters) is rather arbitrary; and each CNN gives a different probability estimate for a given $x$, $y$ pair.
- a huge model like ours is prone to over-fit the values of $\hat{P}(y_i|x_i)$ to the training data.
- the amount of regularization is not precisely controlled.
- we are using the CNN as a component in a bigger model that includes a language prior.

To this end, we consider recalibrating the learned probabilities by further penalizing the weights of the softmax layer.

The class probability for the $k^{th}$ class is given by

$$(6.10) \qquad \hat{P}(k|x) = \frac{e^{\beta_k^T A}}{\sum_{j=1}^{K} e^{\beta_j^T A}},$$

where $A$ is the input to the softmax layer (for a given image $x$) and $\{\beta_k\}$ define the weight matrix $W$ of (5.8). One could scale all the coefficients $\beta_j$ by a fixed quantity $\lambda$ and the predictions of the neural network do not change, only the confidences in them do. Define the biased probabilities $\hat{P}_\lambda(k|x)$ as

$$(6.11) \qquad \hat{P}_\lambda(k|x) = \frac{e^{\lambda \beta_k^T A}}{\sum_{j=1}^{K} e^{\lambda \beta_j^T A}}.$$

Here, multiplication by $\lambda$ scales the norm of the $\beta$ vectors, thus acting as a quadratic penalty. As $\lambda \downarrow 0$, all the classes will tend to an equal probability. And as $\lambda \uparrow \infty$, one class will completely dominate the rest. For the CNN to work well with the $n$-gram model, it is important to multiply the log-probabilities it gives by the correct amount of $\lambda$. Thus $\lambda$ becomes an additional tuning parameter for the combined model.

A higher $\lambda$ will give the classifier more power in the end-to-end system; similarly as smaller $\lambda$ will give the n-gram model more power. $\lambda = 1$ should work well in a scenario where we are processing a text that uses a font seen by the CNN and a written dialect well approximated by the n-gram. If we are processing a page of Sanskrit text written in Telugu script (like 'Rāmayaṅa' in Figure 18), we might want to give the CNN more influence relative to the n-gram. On the other hand, if the font were not seen by the CNN (like the archaic 'Kōśa' and 'Annamayya' samples in Figure 18) we might want to give the n-gram model more influence. (Although a well regularized CNN should be immune to font-drifts.)

**7. Comprehensive Evaluation.** To test the integrity of the three modules working together, we run an evaluation suite on six sample texts. These texts, with varying levels of degradation, conform to the scope of our problem to various levels. The results are summarized in Table 5 and the sample texts are shown in Figure 18.

TABLE 5

*Error-rates of the end-to-end system for the sample texts in Figure 18. The columns are: cross-reference to Figure 18; name of the text; language of the text; whether the CNN is trained on the font in the text, 'Similar' means a similar font has been trained on; number of glyphs in the sample text; number of broken glyphs + number of glyph attachments; error rates without and with the language model(LM) and from Google's OCR system. The best performances are shown in bold.*

|   | Name of the text | Language | Trained Font | Count | Broken + Attached | Errors w/o LM | Errors with LM | Errors by Google |
|---|---|---|---|---|---|---|---|---|
| a. | Nudi | Telugu | Yes | 119 | 01(.8%) | 3(2.5%) | **0(0.0%)** | **0(0.0%)** |
| b. | Kōśa | Telugu | No | 155 | 29(19%) | 12(7.7%) | **4(2.6%)** | 5 (3.2%) |
| c. | Annamayya | Telugu | No | 132 | 11(08%)+1 | 12(9.1%) | **5(3.8%)** | 11 (8.3%) |
| d. | Nannayya | Telugu | Similar | 129 | 33(26%) | 13(9.3%) | **7(5.4%)** | 17(13.2%) |
| e. | Bhārata | Mix | Similar | 125 | 25(20%)+3 | 9(7.2%) | **7(5.6%)** | 13(10.4%) |
| f. | Rāmayaṅa | Sanskrit | No | 129 | 26(20%) | 12(9.3%) | 15(11.6%) | **10(7.8%)** |
|   | Total |   |   | 789 | 125(15%)+4 | 61(7.7%) | **38(4.8%)** | 56(7.1%) |

For our evaluation we use the `traditional` architecture augmented with location information (line 3 in Table 4). It uses 1.15M parameters to give a test error of .81%. For the language model, we learn the trigram probabilities from a modern corpus of 43M Unicode Telugu characters obtained from the internet. We get perfect recovery for the very well printed and scanned 'Nudi' text that uses a modern font. However, it is not a realistic specimen. To better evaluate performance on our target documents, we test on fonts that have not been seen by the neural network. It can be seen from Table 5 that the network is not susceptible to data/font-drift, as the error-rates are not higher for unseen fonts (of 'Kōśa' and 'Annamayya'). Our implementation of the recognition-graph also gives us significant improvement in performance, as it extends the scope of the problem to text with erasure of ink. We recover well from a scenario where as high as one in four glyphs is broken ('Nannayya'). We also see that implementation of the

FIGURE 18. *The six sample texts used in the end-to-end evaluation as summarized in Table 5. The first sample is artificially rendered, while the rest are real-world specimens from scanned printed books.*

$n$-gram model further reduces the error-rates for Telugu texts. However, it introduces more errors for the Sanskrit text 'Rāmayaṅa' that is written in Telugu script. This is understandable given that we use a corpus of modern Telugu to learn the trigram probabilities. One could recalibrate the probability estimates given by the CNN (as described in Section 6.4) according to how similar the scanned text is going to be to modern internet Telugu.

Thanks to Deep Learning, we seem to be doing consistently better than the Google OCR system (Smith, 2007). However, they seem to be doing better than us on the Sanskrit text 'Rāmayaṅa' (for the same reason stated above). All in all, we think our system gives satisfactory performance over diverse looking documents. While this performance might not be enough to replace a human transcriber, it is good enough to enable search facility for a digital corpus of scanned texts.

**8. Conclusion.**   We took up the challenging task of Telugu OCR and have designed and implemented an end-to-end framework to solve it. This system is very generalizable to other Brahmic alphasyllabic scripts like Kannada, Malayalam, Gujarati, Punjabi, etc. — the ones that do not use a headline. Heuristics specific to the Telugu script played an important role in solving the problem. This questions our framework's generalizability. While the heuristics used to detect lines are easily generalizable, the ones used to combine glyphs need to be redesigned. We use the latest advances in deep learning to design a very good classifier that is at the heart of the system. We then use classical techniques like $n$-gram modelling and the Viterbi algorithm to equip the model with a good language prior. Although the scope of the problem was originally restricted to well scanned documents without erasure, we later expand it to handle documents with erasure. Similarly one can extend the model to handle joined glyphs. To do this, one just needs to initially generate candidate splits for the edges of the original linear graph of Figure 15. However, even with this expanded scope, we can not expect to achieve human-level performance. This is because the language models employed by humans are are far more complex (as studied in greater detail in Natural Language Processing).

Our framework generalizes to scripts like Devanagari and Arabic with modifications to the segmentation algorithm. These scripts are written with words (not glyphs) as connected components. Modern Deep Learning techniques like Recurrent Neural Networks with Connectionist Temporal Classification(CTC) (Graves et al., 2009) bypass the segmentation task altogether. They do not need the data to be segmented in a way such that there is a one-to-one correspondence between input samples and output labels. While CTC broadens the scope of the problem, it is more difficult to train. It might also need a more complicated language model. Our framework can be used as a reference for more advanced CTC models of the future.

8.1. *Software.* All the software for the end-to-end system is freely available at github.com/TeluguOCR. The software for data generation, along with the required fonts and corpora are also available there.

## References.

ADDRESS OF THE AUTHORS
DEPARTMENT OF STATISTICS
STANFORD UNIVERSITY
390, SERRA MALL
STANFORD, CA, USA
E-MAIL: rakesha@stanford.edu
        hastie@stanford.edu URL: http://stanford.edu/~hastie